\documentclass[conference,compsoc]{IEEEtran}
\usepackage{amsmath}   % From the American Mathematical Society
\usepackage{cite}
\usepackage{amsfonts,amssymb}
\usepackage{mathtools}
\usepackage{amssymb}
\usepackage{graphics}
\usepackage{subcaption}
\usepackage[demo]{graphicx}
\usepackage{color}
\usepackage{url}
\usepackage{soul}
\usepackage{threeparttable}
\usepackage{algorithmicx}
\usepackage[noend]{algpseudocode}
\usepackage[ruled,linesnumbered,noend]{algorithm2e} % For typesetting algorithms
\ifCLASSINFOpdf

\else

\fi

\begin{document}
\title{A Domain Knowledge \textemdash Enabled Hybrid Semi-Supervision Learning Method}
%\author{\IEEEauthorblockN{Yifu Wu and Jin~Wei\\}
%    \IEEEauthorblockA{Department of Electrical \& Computer Engineering\\
%     The University of Akron, Akron, OH 44325, USA} }

\author{  
\IEEEauthorblockN{Yifu Wu}
\IEEEauthorblockA{Department of Electrical and\\
Computer Engineering\\
University of Akron\\
Akron, Ohio 44325-3904\\
Email: yw62@zips.uakron.edu}
    
\and
\IEEEauthorblockN{Jin Wei}
\IEEEauthorblockA{Department of Electrical and\\
Computer Engineering\\
University of Akron\\
Akron, Ohio 44325-3904\\
Email: jwei1@uakron.edu}

\and
\IEEEauthorblockN{Rigoberto Roche}
\IEEEauthorblockA{NASA Glenn Research Center\\
Cleveland, Ohio\\
Email: rigoberto.roche@nasa.gov}
} 

% make the title area
\maketitle
\begin{abstract}
Due to the advances in dramatically increased computing power, machine learning technologies have been widely exploited to provide promising solutions in different application fields. However, the high performance of many of these machine learning technologies highly relies on the availability of sufficient annotated data. Considering the fact that data annotation is an extremely time-consuming process, this condition is not always practical. To address this challenge, in this paper we propose a novel computing method, called hybrid semi-supervision machine learning, that exploits the domain knowledge to enable the accurate results even in the presence of limited labeled data. Simulations results illustrate the effectiveness of our method.
\end{abstract}

\IEEEpeerreviewmaketitle

\section{Introduction}\label{sec:introduction}
In the past ten years, machine learning, especially deep learning, achieves an extensive success in many tasks of supervised learning. Achieving an efficient predictive model of supervised learning always requires adequate annotated training data, which is not always feasible considering the fact that data annotation is an extremely time-consuming process. Therefore, it is desirable to develop a machine learning method that enables satisfactory accuracy in the presence of limited labeled data, which is called weakly-supervised learning.

Weakly-supervised machine learning can be classified into three main types, incomplete supervision, inexact supervision, and inaccurate supervision~\cite{zhou2017brief}, of which incomplete supervision is the most widely used one. Generally speaking, incomplete supervision assumes that training data are mostly unlabeled and only a small subset of them is annotated. To analyze the unlabeled data, there are two main types of approaches: (1) active learning, which utilizes the knowledge of domain experts to selectively label a small amount of training data valuable for training model~\cite{settles2012active}, and (2) semi-supervision learning that integrates the supervised learning and unsupervised learning. There are various semi-supervision learning algorithms have been developed by exploiting generative model, low-density separation, graph-based model, or heuristic model to analyze the unlabeled data~\cite{chapelle2009semi,lee2013pseudo,rasmus2015semi,laine2016temporal,tarvainen2017mean}. These methods are effective in the considered scenarios. However, their performances still rely on the number of labeled data. On the other hand, the domain knowledge has been proved useful to regularize the posterior\cite{ganchev2010posterior}\cite{hu2016harnessing}\cite{hu2016harnessing}\cite{stewart2017label}\cite{hu2018deep}. However, these existing works mainly focus on regularizing the posterior of the model output, and thus require dedicated constraint design before model training, which is not always practical in the tasks with high uncertainty. To address these challenges, in this task we propose a novel hybrid semi-supervision learning method whose essential idea is to explore the domain knowledge effectively and to integrate the features characterized by the domain knowledge and the features presented by the labeled and unlabelled data. The authors would like to claim that the framework presented in this paper has been included in a provisional patent~\cite{patent}. 

In the next section, we discuss the related works about heuristic learning and constraint learning for semi supervision learning. Section~\ref{sec:model} illustrates the details of our proposed domain knowledge-enabled hybrid semi-supervision learning method. The simulation results and conclusions are presented in Sections~\ref{sec:simulation}~and~\ref{sec:conclusion}, respectively.

\section{Related Work}\label{sec:related_work}
\subsection{Heuristic Models for Semi-Supervision Learning}
In many established semi-supervision learning techniques, heuristic models have been used with a prerequisite that the labeled and unlabeled data have similar distributions. The authors in \cite{lee2013pseudo} proposed a bootstrapping method by utilizing a neural network inference of the unlabeled data to generate pseudo labels that are used to regularize the neural network in turn. In this work, the cost function of the neural network is defined based on cross entropy as follows: 
\begin{equation}\label{eqn:pseudo}
\begin{split}
L_{p}=&\frac{1}{n}\sum_{m=1}^{n}\sum_{i=1}^{C}E(y_{i}^{m}, f_{i}^{m})\\
&+\alpha (t)\frac{1}{n'}\sum_{m=1}^{n'}\sum_{i=1}^{C}E({y'}_{i}^{m}, {f'}_{i}^{m})
\end{split}
\end{equation}
where $n$ is the number of labeled training data per mini-batch if stochastic gradient descent (SGD) is applied, $n'$ is the number of unlabeled data per mini-batch, $C$ is the number of classes, $E(y_{i}^{m}, f_{i}^{m})$ is the cross-entropy between the true label and inference result for the $m$th labeled data, and $E({y'}_{i}^{m}, {f'}_{i}^{m})$ is the cross-entropy between the pseudo label and inference result for the $m$th unlabeled data for Class~$i$. Since both the pseudo and true labels are one-hot vectors, $y_{i}^{m}$ and ${y'}_{i}^{m}$ are both binary. $\alpha (t)$ is a ramp-up function to determine the contribution of the unlabeled data for the overall learning result. Since the inference result for the unlabeled data has high uncertainty in the initial stage of training, $\alpha (t)$ is initialized as $0$ and gradually increases as the neural network learns more knowledge from the labeled data.  

The first part of the loss function formulated in Eq.~(\ref{eqn:pseudo}) is the loss for the supervised learning based on the labeled data and the second part in Eq.~(\ref{eqn:pseudo}) is the loss for the unsupervised learning based on the unlabeled data. There is one essential challenge for using this method. The pseudo labels, whose quality is determined by the effectiveness of the neural network, eventually impact the efficiency of the neural network. In other words, if wrong pseudo labels are incorporated with the labeled data during the training process, the model error of the neural network will be exacerbated as the ramp-up index $\alpha (t)$ increases. To enhance the robustness of the prediction of pseudo labels, the semi-supervision techniques in~\cite{rasmus2015semi,laine2016temporal,tarvainen2017mean} apply self-ensembling training to develop multiple child models with diverse variants from a parent model. Generally speaking, data augmentation and diverse configurations of neural networks are two main approaches to generate various child models. Although child models have diverse variants, their outputs are expected to be consistent with each other. To smooth the prediction of pseudo labels of the unlabeled data, ensemble learning proposed in~\cite{bachman2014learning} is widely adopted for fusing the child models. 

\subsection{Domain Constraints for Semi-Supervision Learning}
To address the challenge of limited annotated data, some semi-supervision learning techniques exploit domain knowledge to generate structured functions such as structural constraints\cite{ganchev2010posterior} or logic rules\cite{hu2016harnessing}. These structured functions provide a complement way to regularize the distribution of model posterior. In these developed methods, the set of distribution of domain constraints with respect to the model posterior is formulated as: \begin{equation}\label{eqn:domain_constraints}
\begin{split}
% \textit{Posterior Constraints Set: }
Q=\{q(
\mathbf{y}): \mathbb{E}_{q}\left[\mathbf{G}\right]\leq \mathbf{c}\}
\end{split}
\end{equation}where $\mathbf{G}=\left\{G_i(\mathbf{x},\mathbf{y})\right\}$ is a set of constraint functions and its expectation is bounded by $\mathbf{c}$. The constraint set implies extra information on posterior distribution and narrows down the searching space of posterior. To enforce the posterior distribution to the desired implicit distributions according to the domain knowledge constraints, a penalty term is defined to measure the distances between these two distributions by using Kullback-Leibler (KL) divergence as follows: \begin{equation}\label{eqn:kl_domain_constraints}
\begin{split}
D_{KL}\left(Q||p(\mathbf{x},\mathbf{y})\right)= \underset{q\in Q}{\mathrm{min}} D_{KL}\left(q(\mathbf{y})||p(\mathbf{x},\mathbf{y})\right)
\end{split}
\end{equation} These domain knowledge-based methods highly rely on the precise information of domain knowledge, which requires the explicitly stated feature set $\mathbf{G}$ prior to the neural network training and thus limits the applications of semi-supervision learning techniques. Additionally, $\mathbf{G}$ is only determined by the input $\mathbf{x}$ and output $\mathbf{y}$ of the targeted task model, which is not always the situation since many task models, such as physical systems, normally also relate to the system parameters besides $\mathbf{x}$ and $\mathbf{y}$.

\section{Our Proposed Hybrid Semi-Supervision Learning Method}\label{sec:model}
\begin{figure*}[!ht]
\centering
\includegraphics[height=4cm,width=4.8in]{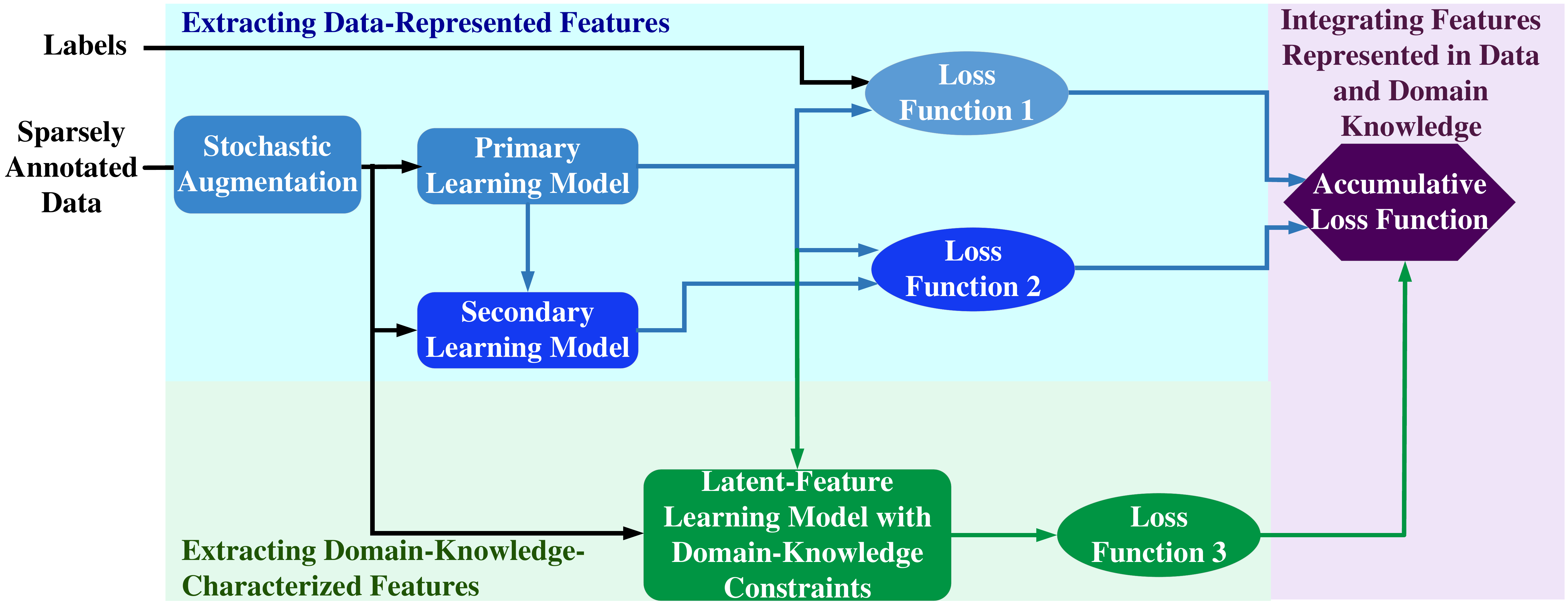}
\caption{Overview of our proposed knowledge domain\textemdash enabled hybrid semi-supervision learning method.}
\label{Fig:weak_supervision}
\end{figure*}

The overview of our proposed knowledge domain\textemdash enabled hybrid semi-supervision learning method is shown in Fig.~\ref{Fig:weak_supervision}. As shown in Fig.~\ref{Fig:weak_supervision}, our proposed method mainly comprises primary, secondary, and latent-feature neural network (NN)-based learning models. In the following content, we use a data-driven $C$-class classification problem to explain the details of our proposed learning method. 

The primary learning model $f_{\theta_p}\left(\cdot\right)$, where $\theta_p$ represents the parameters of the primary learning model, is designed to extract the features from the annotated data and to interact with the other two learning models. To achieve this goal, the Loss Function $L_1\left(\theta_p\right)$ is formulated to minimize the total distance between the true labels $y_i^m\in\mathbf{y}$ and the corresponding inference results $f_{\theta_p}^{i}\left(x_m\right)$ for labeled data $x_m\in\mathbf{x}$ achieved by the primary learning model as follows:\begin{eqnarray}
\label{loss1}
L_{1}\left(\theta_p\right)=\frac{1}{N}\sum_{m=1}^{N}\sum_{i=1}^{C}E_{\theta_p}\left(y_i^m,f_{\theta_p}^{i}\left(x_m\right)\right)
\end{eqnarray} where $N$ is the number of annotated data and $C$ is the number of classes. $E_{\theta_p}\left(\cdot,\cdot\right)$ is the cross-entropy between the true labels and the corresponding inference results. Additionally, stochastic data augmentation is applied to increase the effective size of existing labeled data. 

The secondary learning model $f_{\theta_s}\left(\cdot\right)$, where $\theta_s$ denotes the parameters of the secondary learning model, is designed to interact with the primary learning model to extract the features from the raw data by obtaining the pseudo labels of these unlabelled data. To achieve this goal, firstly, the secondary learning model is constructed based on the structure of the primary learning model. In our work, the model parameters $\theta_s$ of the secondary learning model are achieved by calculating the exponential moving average of the historical values of $\theta_p$~\cite{tarvainen2017mean}, $\theta_{s,t}\leftarrow\beta\theta_{s,t-1}+\left(1-\beta\right)\theta_{p,t-1}$ where $\beta$ is a smoothing hyperparameter and $t$ is the index of the current epoch or step. Secondly, the interaction between the primary and secondary learning models via Loss Function $L_2\left(\theta_p,\theta_s\right)$ that is formulated to fuse the features extracted by using the primary and secondary learning models and to maximize the consistency between the prediction of the primary and secondary learning models as follows:\begin{eqnarray}
L_{2}\left(\theta_p,\theta_s\right)=\alpha(t)\left[\frac{1}{N+N'}\sum_{m=1}^{N+N'}\sum_{i=1}^{C}D_{KL}\left(f_{\theta_s}^{i}\left(x_m\right)||f_{\theta_p}^{i}\left(x_m\right)\right)\right]\nonumber
\end{eqnarray}
%\vspace{-0.5cm}
\begin{eqnarray}
\label{loss2}
\end{eqnarray}where $\alpha(t)$ is a ramp-up function to determine the contribution of the secondary learning model, which is initialized as $0$ and increases gradually, and $N'$ is the number of raw data. Furthermore, to enhance the generalization of the interaction between the primary and secondary learning models and to mitigate the uncertainty introduced by the scarcity of the labeled data, we realize dynamic model reconfiguration of the secondary learning model by applying the multiplicative noise techniques~\cite{Neftci17}, Dropout and DropConnect algorithms~\cite{srivastava2014dropout,wan2013regularization}. 

The latent-feature learning model $f_{\theta_l}\left(\cdot\right)$, where $\theta_l$ denotes the parameters of the latent-feature learning model, is designed to exploit the domain knowledge for the task model to regularize the latent features $\ell$ extracted by the primary learning model $f_{\theta_p}\left(\cdot\right)$. The latent features $\ell$ can be the soft prediction vector $f_{\theta_p}(\mathbf{x})$ or the output of the last hidden layer of the primary model $f_{\theta'_p}(\mathbf{x})$ where $\theta'_p$ denote the weights of the primary learning model except those of the output layer. Considering that $f_{\theta'_p}(\mathbf{x})$ exhibits more feature information than $f_{\theta_p}(\mathbf{x})$, in our work we select $\ell=f_{\theta'_p}(\mathbf{x})$. However, the latent features $\ell$ are abstract and not directly related to the domain knowledge. To address this challenge, the latent-feature learning model is developed to transfer the latent features $\ell$ to the approximate unknown critical-parameter vector $\mathbf{z}$ of the targeted task, such as the unknown critical parameters of a physical system for the task. In other words, $f_{\theta_l}(\ell) \approx \mathbf{z}$. The latent-feature learning model is trained to gradually optimize the approximated critical parameters $\mathbf{z}$ within a parameter space restricted by the domain knowledge, which results in regularizing the latent features $\ell$,  the conditional probability $p_{\theta'_p}(\ell|x)$,  and eventually the primary learning model $f_{\theta_p}\left(\cdot\right)$. Let $\tilde{\mathbf{G}}=\left\{\tilde{G}_i\left(\mathbf{x},f_{\theta_l}(\ell), \mathbf{z}'\right)\right\}$ be a domain knowledge constraint set, where $\mathbf{z}'$ denotes the available parameter vector. The domain knowledge constraints we consider in our current semi-supervision learning model is modeled as follows:\begin{equation}\label{eqn:domain_knowledge}
\begin{split}
\tilde{G}_i(\mathbf{x},f_{\theta_l}(\ell),\mathbf{z}') \leq c_i,
\end{split}
\end{equation}where $c_i\in\mathbf{c}$ is a boundary parameter for the domain knowledge constraint $\tilde{G}_i(\cdot)$. The Loss Function $L_3\left(\theta'_p,\theta_l\right)$ is formulated as follows: \begin{eqnarray}
\label{loss3}
L_{3}\left(\theta'_p,\theta_l\right)=D_{KL}\left(q_{\theta_{l}}\left(\ell\right)\|p_{\theta'_{p}}\left(\ell|\mathbf{x}\right)\right)+\nonumber\\ \gamma\big\|\mathbb{E}_{\theta_l}\left[\tilde{\mathbf{G}}\left(\mathbf{x},f_{\theta_l}(\ell),\mathbf{z}'\right)\right]-\mathbf{c}\big\|_{\beta}
\end{eqnarray}where $q_{\theta_{l}}(\ell)$ is an auxiliary variational probability, which is learned via posterior regularization according to the domain knowledge constraints $\mathbf{G}\left(\cdot\right)$, and is optimized by imitating the conditional probability $p_{\theta'_{p}}\left(\ell|\mathbf{x}\right)$. The first KL-divergence term in Eq.~(\ref{loss3}) is formulated to enforce the model posterior $p_{\theta'_{p}}\left(\ell|\mathbf{x}\right)$ to approach the desired distribution space $q_{\theta_{l}}\left(\ell\right)$ based on the domain knowledge. The second term (a norm $||\cdot||_{\beta}$) denotes the penalty cost of bounded domain knowledge constraints. $\gamma$ is set to adjust the weight of the second term. By using Eqs.(\ref{loss1})~to~(\ref{loss3}), the accumulative loss function in Fig.~\ref{Fig:weak_supervision} can be calculated as $L=L_1+L_2+L_3$, which is used to train our semi-supervision learning model. 

The training procedure of our proposed hybrid semi-supervision learning algorithm is implemented based on Expectation Maximization (EM) algorithm~\cite{ganchev2010posterior, hu2016harnessing}. In addition, we utilize a self-ensembling approach to generate pseudo labels for unlabeled data to reinforce the generalization of the primary learning model. The details of the training procedure are illustrated in Algorithm~\ref{training_1}. Step $3$ in Algorithm~\ref{training_1} is similar to the E-step of posterior regularization. However, in our method, it is designed to regularize the critical parameters for the task model, such as the critical parameters of a physical system for the task, which are not limited to the target output $\mathbf{y}$. Step $4$ is similar to the M-step of posterior regularization, which refers to the updated desired distribution of posteriors and optimizes the weights of primary network $\theta_p$. Steps $5$ and $6$ are the steps to ensemble the primary and secondary learning models.
\begin{algorithm}
    Initialize the parameters of the primary, secondary, and feature-latent neural networks, $\theta_p$, $\theta'_p$ $\theta_s$ and $\theta_l$ where $\theta'_p\subsetneq\theta_p$\;
    \While{not converging}{
        $\theta^{t+1}_{l}\gets\underset{\theta_l}{\mathrm{argmin}}L_3\left({\theta'}^{t}_{p}, \theta_{l}^{t}\right)$\;
        $\theta_{p}^{t+1}\gets\underset{\theta_p}{\mathrm{argmin}}L_3({\theta'}_p^{t}, \theta_{l}^{t+1})$\;
        $\theta_{p}^{t+1}\gets\underset{\theta_p}{\mathrm{argmin}}(L_1\left(\theta_{p}^{t+1}\right)+L_2\left(\theta_{p}^{t+1},\theta_{s}^{t+1}\right))$\;
        $\theta_{s}^{t+1}\gets\beta \theta_p^{t+1} + (1-\beta)\theta^{t}_p$
    }
    Done\;
    \caption{Domain Knowledge\textemdash Enabled Hybrid Semi-Supervision Learning Method}
    \label{training_1}
\end{algorithm}

\begin{figure*}[!ht]
\centering
\includegraphics[height=5.2cm,width=5in]{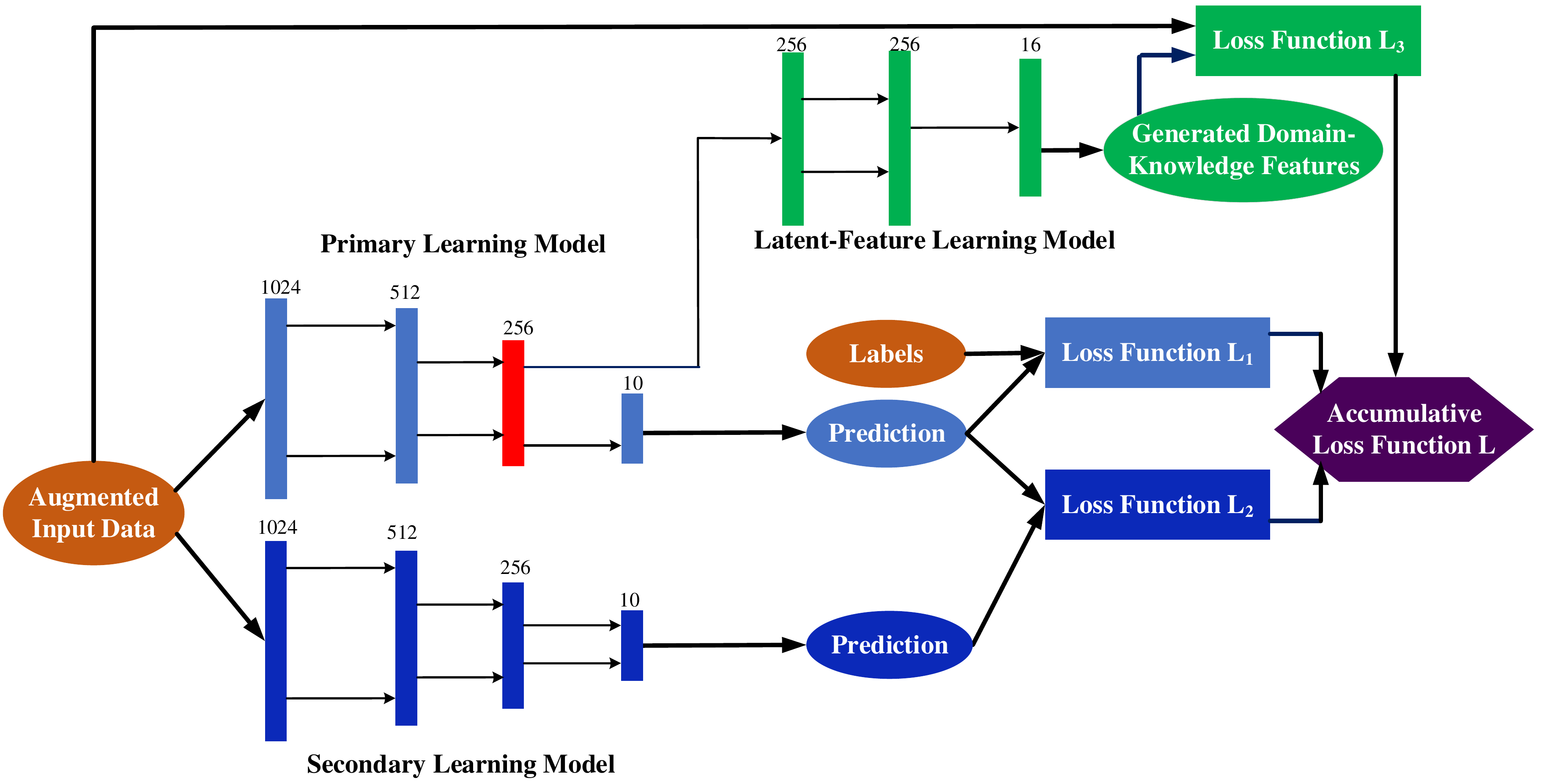}
\caption{The specific structure of our hybrid semi-supervision learning method.}
\label{Fig:power_network}
\end{figure*}

\begin{table*}[ht]
\caption{Performance comparison between our method and a established semi-supervision learning method.} % title of Table
\centering
\begin{tabular}{| c | c | c | c | c | c |}
\hline  \# of Labels & 90(1.25\%) & 180(2.5\%) & 360(5\%) & 720(10\%) \\\hline 
Established Semi-Supervision Method & $90.47\pm 1.89$ & $96.64\pm 3.28$ & $99.90\pm 0.10$ & $99.95\pm 0.05$\\\hline 
Our Domain Knowledge\textemdash Enabled Semi-Supervision Method & $99.74\pm 0.29$ & $99.89\pm 0.09$ & $99.96\pm 0.04$ & $99.98\pm 0.03$\\\hline 
\end{tabular}
\label{Table:power_result}
\end{table*}

\section{Simulation Results}\label{sec:simulation}

In this section, we evaluate the performance of our proposed knowledge domain\textemdash enabled hybrid semi-supervision learning method by applying it to detect the location of the power outage attack in a WSCC 9-Bus Power System. The details of the power system can be found in~\cite{Al-Hinai00}. In this scenario, the frequencies of the three power generators are recorded and used as the training data for the attack detection. The specific structure of our hybrid semi-supervision method deployed in this scenario is shown in Fig.~\ref{Fig:power_network}. Comparing the specific implementation shown in Fig~\ref{Fig:power_network} and the general structure illustrated in Fig.~\ref{Fig:weak_supervision}, it can be seen that the primary learning model is realized via a four-layer dense neural network, the secondary learning model has a similar structure as that of the primary model, and the latent-feature learning model is executed as a three-layer dense neural network. The latent features, which are the output of the last hidden layer of the primary learning model, are used as the input of the latent-feature learning model. 

\begin{figure}[!ht]
\center {\includegraphics[width=0.40\textwidth]{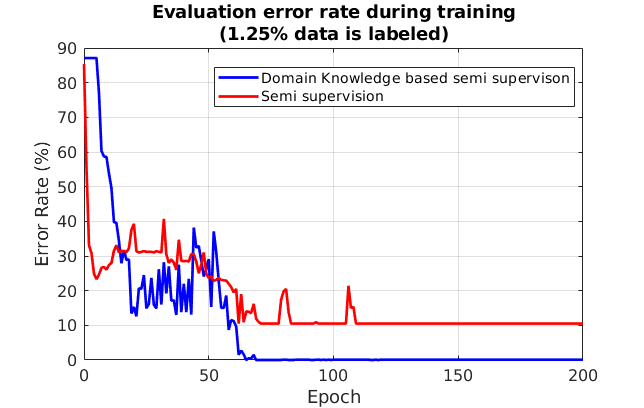}}
\caption{\label{Fig:power_result} Validation accuracy during training process}
%\vspace*{-0.4cm}
\end{figure}

In this scenario, we apply the transient stability constraints to formulate the domain knowledge constraint set $\tilde{\mathbf{G}}$ in our hybrid semi-supervision learning method, which includes the swing equation that is a second-order differential equation modeling the physical coupling between the synchronous generators in the power system and defined as follows~\cite{Grainger94}: \begin{eqnarray}\label{eqn:swing_equation}
M_{i}\overset{\cdot}{\omega}_{i}(t)=-D_{i}\omega_{t}(t)+P_{mi}(t)-|E_{i}|^{2}G_{ii}(t)\nonumber \\
-\sum_{j=1,j\neq i}^{N}P_{ij}(t)\sin\big[\theta_{i}(t)-\theta_{j}(t)+\varphi_{ij}(t)\big],
\end{eqnarray}where
\begin{eqnarray*}\label{eqn:swing_equation_addition}
\left\{
  \begin{array}{ll}
    \theta_{i}(t)=\theta_{c}(t)+\int_{0}^{t}\omega_{t}(\tau)d\tau,\\
    \varphi_{ij}(t)=\arctan\left[\frac{G_{ij}(t)}{B_{ij}(t)}\right].
  \end{array}
\right.
\end{eqnarray*}where $E(t)$, $P(t)$, $G(t)$, and $B(t)$ denote the internal voltage, mechanical power input, Kron-reduced equivalent conductance, and Kron-reduced equivalent susceptance, respectively. These critical system parameters are determined by the topology of the power system. Due to the lack of information on the power outages, such as locations, the values of these parameters are unknown. The latent-feature learning model in our semi-supervision learning method transfers the latent features, which are the output of the last hidden layer of the primary learning model, to approximate the values of these parameters and to optimize the values gradually within a parameter space restricted by Eq.~(\ref{eqn:swing_equation}), which results in regularizing the primary learning model. Besides the swing equation in Eq.~(\ref{eqn:swing_equation}), the frequency synchronization and phase angle cohesiveness of synchronous generators for transient stability~\cite{Grainger94} are also included in the domain knowledge constraint set via logic AND and OR functions. By implementing our semi-supervision learning method via the Algorithm~\ref{training_1}, we obtain the simulation results in Table~\ref{Table:power_result}, which shows the accuracy in detecting the locations of power outage attacks in the presence of different sparsity levels of annotated data. As shown in Table~\ref{Table:power_result}, we also compare the performance of our proposed method with that of an established bootstrapping-based semi-supervision learning method proposed in~\cite{tarvainen2017mean}. From Table~\ref{Table:power_result}, we can observe that when there are at or above $5$~\% of the data are labeled, our proposed method achieves comparable performance with the established method. When the percentage of the labeled data is $\leq 2.5~\%$, our hybrid semi-supervision learning method outperforms the established method. To illustrate the performance of our proposed method in more detail, in Fig.~\ref{Fig:power_result} we show the evaluation error-rates of our method (in blue) and the established semi-supervision method (in red) versus training epochs when there are $1.25$~\% data annotated. From Fig.~\ref{Fig:power_result}, we can observe that after $50$ training epochs, our method converges to lower error rate faster than the established method. 

\section{Conclusions}\label{sec:conclusion}
In this paper, we propose a novel computing method, called hybrid semi-supervision machine learning, that can achieve high performance even in the presence of limited annotated data. The essential idea of our proposed method is to explore the domain
knowledge efficiently and to integrate the features characterized by the domain knowledge and the features presented by the labeled and unlabelled data. The simulation results illustrate the effectiveness of our method by comparing the performance of our method with an established bootstrapping-based semi-supervision method. In our ongoing work, we are working on evaluating the performance of our proposed method in different scenarios and improving our proposed method from the perspective of characterizing more complex domain knowledge constraints. 
\section{ACKNOWLEDGEMENTS}\label{sec:acknowledgement}
This work was supported by an Early Career Faculty grant from NASAs Space Technology Research Grants Program. 
\bibliographystyle{ieeetr}
\bibliography{IEEEbibfile}

\end{document}